\documentclass[letterpaper, 10 pt, conference]{ieeeconf}  
\IEEEoverridecommandlockouts
\usepackage{cite}
\usepackage{amssymb}
\usepackage{amsmath,amssymb,amsfonts}

\usepackage{graphicx}
\usepackage{textcomp}
\usepackage{xcolor}
\usepackage{listings}
\usepackage[english]{babel}
\usepackage[utf8]{inputenc}
\usepackage{algorithm}
\usepackage{algorithmic}
\usepackage{multirow}
\usepackage{amsfonts}
\usepackage{mathtools}
\usepackage{flushend}

\usepackage{url}
\usepackage{hyperref}
\makeatletter
\newcommand{\printfnsymbol}[1]{%
  \textsuperscript{\@fnsymbol{#1}}%
}
\makeatother

\def\BibTeX{{\rm B\kern-.05em{\sc i\kern-.025em b}\kern-.08em
    T\kern-.1667em\lower.7ex\hbox{E}\kern-.125emX}}
\begin{document}

\title{Robotic Control Using Model Based Meta Adaption}

\author{ Karam Daaboul$^{*}$, Joel Ikels$^{*}$ and J. Marius Z\"ollner
\thanks{*Equal contributions}
\thanks{Karlsruhe Institute of Technology, Kaiserstr. 12, 76131 Karlsruhe, Germany
        {\tt\small \{daaboul, marius.zoellner\}@kit.edu, joel.ikels@student.kit.edu}}%
}
\maketitle
\begin{abstract}
In machine learning, meta-learning methods aim for fast adaptability to unknown tasks using prior knowledge. Model-based meta-reinforcement learning combines reinforcement learning via world models with Meta Reinforcement Learning (MRL) for increased sample efficiency. However, adaption to unknown tasks does not always result in preferable agent behavior. This paper introduces a new Meta Adaptation Controller (MAC) that employs MRL to apply a preferred robot behavior from one task to many similar tasks. To do this, MAC aims to find actions an agent has to take in a new task to reach a similar outcome as in a learned task. As a result, the agent will adapt quickly to the change in the dynamic and behave appropriately without the need to construct a reward function that enforces the preferred behavior.

\end{abstract}

\section{Introduction}
\begin{figure}
\centerline{\includegraphics[scale=0.16]{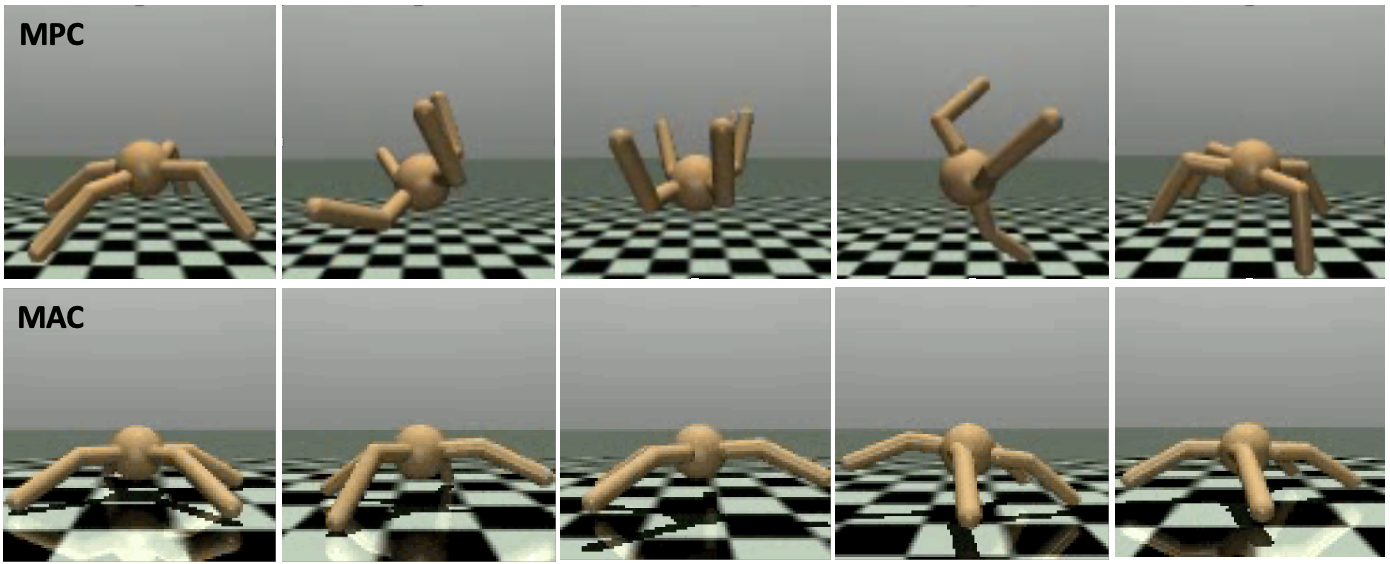}}
\caption{Action sequences of an ant robot during meta-testing. The test task is to adapt to the gravity of $5 \: m/s^2$. A model-based meta-reinforcement learning approach with MPC results in an undesired robot behavior (row 1). MAC finds a behavior similar to the one learned at its reference task (row 2).}
\label{ant_roll}
\end{figure}

Adaptive behavior lies in the very nature of life as we know it. By forming a variety of behaviors, the animal brain enables its host to adapt to environmental changes continuously \cite{sterling2015principles}. Toddlers, for example, can learn how to walk in the sand in several moments, whereas robots often struggle to adapt fast and show rigid behavior encountering a task not seen before. Fast adaption is possible because animals do not learn from scratch and leverage prior knowledge to solve a new task. In machine learning, the domain of meta-learning takes inspiration from this phenomenon by enabling a learning machine to develop a hypothesis on how to solve a new task using information from prior hypotheses of similar tasks \cite{schmidhuber:1987:srl}. Thus, it aims to learn models that are quickly adaptable to new tasks and can be described as a set of methods that apply a learned prior of common task structure to make a generalized inference with small amounts of data \cite{schmidhuber:1987:srl}, \cite{Finn2018}.\newline
The domain of \emph{model-based reinforcement learning} (MBRL) comprises methods that enable a Reinforcement Learning (RL) agent to successfully master complex behaviors using a deep neural network as a model of a tasks system dynamics \cite{Atkeson1997}. To solve an RL task, this dynamics model is utilized to optimize a sequence of actions (e.g., with model predictive control) or to optimize a policy, making MBRL more sample efficient than model-free reinforcement learning (MFRL) \cite{Williams}, \cite{Deisenroth2011}, \cite{Nagabandi2017}. Even though MBRL methods show improved sample efficiency compared to MFRL approaches, the amount of training data needed to reach "good" performance scales exponentially with the dimensionality of the input state-action space of the dynamics model \cite{Chatzilygeroudis2018}. Additionally, data scarcity is even more challenging when a system has to adapt online while executing a task. A robot, for example, might encounter sudden changes in system dynamics (e.g., damaged joints) or changes in environmental dynamics (e.g., new terrain conditions) that require fast online adaption. By combining meta-learning and MBRL, robots can learn how to quickly form new behaviors when the environment- or system-dynamics change \cite{Nagabandi2018}, \cite{Saemundsson2018}, \cite{Kaushik2020}, \cite{Belkhale2021}. However, newly formed behavior might be undesirable even if the underlying task is mastered correctly according to the environment's reward function. For example, as seen in figure \ref{ant_roll}, a robot Ant, trained to walk as fast as possible, will start to jump or roll if the gravity of its environment is very low. In a real-world setting, such a situation might damage the robot. Therefore, the RL agent requires a tailored reward function to form behavior that does no damage. Nevertheless, designing a reward function is challenging since it is time-consuming and challenging to master, especially for various tasks. 
This paper introduces a controller for MBRL that employs meta-learning to apply a selected robot behavior from one robotic task to a range of similar tasks. In other words, it aims to find actions the robot has to take in a new task to reach a similar outcome as in a learned task. Thus, it alleviates the need to construct a reward function that enforces preferred behavior. It builds on top of the FAMLE algorithm by Kaushik et al. \cite{Kaushik2020} making use of an Embedding Neural Network (ENN) for quick adaption to new tasks through task embeddings as learned priors. By combining an ENN with an RL policy of a reference task, the controller predicts which actions need to be taken in unseen tasks to mimic the behavior of the reference task. While being initialized with the most likely embedding, a trained meta is adapted to approximate future environment states and compare them to the preferred states of the reference task. Actions leading to states in the unseen task that are very similar to those reached by the RL policy in the reference task are then chosen to be executed in the environment. To account for the usage and adaption of a meta-model during planning, we call our approach Meta Adaptation Controller (MAC). \newline
First, we introduce related work and preliminaries. Next, the challenge and our approach to solving it are described. Finally, experiment results are presented that compare MAC with MPC employing different meta-learning methods.
\section{Related Work}
In recent years, robotics has achieved remarkable success with model-based RL approaches \cite{Zhang2018},\cite{Nagabandi2019},\cite{yang2020}. The agent can choose optimal actions by utilizing the experiences generated by the model\cite{Nagabandi2017}. As a result, the amount of data required for model-based methods is typically much smaller than their model-free counterparts, making these algorithms more attractive for robotic applications. One drawback in many of these works is the assumption that the environment is stationary. In real robot applications, however, many uncertainties are difficult to model or predict, some of which are internal (e.g., malfunctions \cite{Nagabandi2018}) and others external (e.g., wind{\cite{Belkhale2021}). These uncertainties make the stationary assumption impractical. That can lead to suboptimal behavior or even catastrophic failure. Therefore, a quick adaptation of the learned model is critical. \newline 
"Gradient-based meta-learning methods leverage gradient descent to learn the commonalities among various tasks" \cite[p. 1]{Lee2018}. 
One such method introduced by Finn et al. \cite{Finn2017} is \emph{Model-Agnostic Meta-Learning} (MAML). The key idea of MAML is to tune a model's initial parameters such that the model has maximal performance on a new task. Here, meta-learning is achieved with bi-level optimization, a models task-specific optimization and a task-agnostic meta optimization. Instantiated for MFRL, MAML uses policy gradients of a neural network model, whereas, in MBRL, MAML is used to train a dynamics model. REPTILE by Nicol et al. \cite{Nichol} is the first-order implementation of MAML. In contrast to MAML, task-specific gradients do not need to be differentiated through the optimization process. This makes REPTILE more computationally efficient with similar performance.\newline
A model-based approach using gradient-based MRL was presented in the work of Nagabandi et al. \cite{Nagabandi2018} and targets online adaption of a robotic system that encounters different system dynamics in real-world environments. In this context, Kaushik et al. \cite{Kaushik2020} point out that in an MRL setup where situations do not possess strong global similarity, finding a single set of initial parameters is often not sufficient to learn quickly. One potential solution would be to find several initial sets of model parameters during meta-training and, when encountering a new task, use the most similar one so that an agent can adapt through several gradient steps. Their work \emph{Fast Adaptation through Meta-Learning Embeddings} (FAMLE) approaches this solution by extending a dynamical models input with a learnable d-dimensional vector describing a task. Similarly, Belkhale et al. \cite{Belkhale2021} introduce a meta-learning approach that enables a quadcopter to adapt online to various physical properties of payloads (e.g., mass, tether length) using variational inference. Intuitively each payload causes different system dynamics and therefore defines a task to be learned. Since it is unlikely to accurately model such dynamics by hand and it is not realistic to know every payloads properties value beforehand, the meta-learning goal is the rapid adaption to unknown payloads without prior knowledge of the payload's physical properties. That is why a probabilistic encoder network finds a task-specific latent vector fed into a dynamics network as an auxiliary network. Using the latent vector, the dynamics network learns to model the factors of variation that affect the payload's dynamics and are not present in the current state. All these algorithms use MPC during online adaption. Our work introduces a new controller for online adaption in a model-based meta-reinforcement learning setting.
\section{Preliminaries}

\subsection{Meta Learning}
Quick online adaption to new tasks can be viewed in the light of a few-shot learning setting where the goal of meta-learning is to adapt a model $f_{\theta}$ to an unseen task $\mathcal{M}_{j}$ of a task distribution $p(\mathcal{M})$ with a small amount of $k$ data samples \cite{Finn2017}. The meta-learning procedure usually is divided into meta-training with $n$ meta-learning tasks $\mathcal{M}_{i}$ and meta-testing with $y$ meta-test tasks $\mathcal{M}_{j}$ both drawn from $p(\mathcal{M})$ without replacement \cite{Finn2018}. During meta-training, task data may be split into train and test sets usually representing $k$ data points of a task  $\mathcal{D}_{}^{\text{meta-train}}=\{(\mathcal{D}_{i=1}^{\text{tr}},\mathcal{D}_{i=1}^{\text{ts}}),\ldots (\mathcal{D}_{i=n}^{\mathrm{tr}}, \mathcal{D}_{i=n}^{t s})\}$. Meta-testing task data $\mathcal{D}_{}^{\text{meta-test}}={(\mathcal{D}_{j=1}^{\operatorname{meta-test}}, \ldots, \mathcal{D}_{j=y}^{\operatorname{meta-test}})}$ is hold out during meta-training \cite{Finn2018}. Meta-training is then performed with $\mathcal{D}_{}^{\text{meta-train}}$ and can be viewed as bi-level learning of model parameters \cite{Rajeswaran2019}. In the inner-level, an update algorithm $Alg$ with hyperparameters $\psi$ must find task-specific parameters $\phi_i$ by adjusting meta-parameters $\theta$. In the outer-level, $\theta$ must be adjusted to minimize the cumulative loss of all $\phi_i$ across all learning tasks by finding common characteristics of different tasks through meta parameters $\theta^{\star}$:
\begin{equation}
  \label{eqn:meta-learning-objective}
	\begin{split}
	\overbrace{\theta^{\star}= \arg \min _{\theta}
    \sum_{i=1}^{n}\mathcal{L}_{\mathcal{D}_{i}\sim\mathcal{M}_{i}} (\phi_{i})}^{\text {outer-level }}\\\
    \text { where } \underbrace{{{\phi_{i}}}=Alg^{\psi}_{\mathcal{D}_{i}\sim\mathcal{M}_{i}}(\theta)}_{\text{inner-level}}
	\end{split}
\end{equation}

Once $\theta^{\star}$ is found, it can be used during meta-testing for quick adaption:
${{\phi_{j}}}=Alg(\theta^{\star},\mathcal{D}_{j})$

\subsection{Model-based Reinforcement Learning}
In RL, a task can be described as a Markov Decision Process (MDP) $
\mathcal{M}=\left\{ S, A, p\left({s}_{t=0}\right), p\left({s}_{t+1} \mid {s}_{t},{a}_{t}\right),r, H\right.\}
$ with a set of states $S$, a set of actions $A$, a reward function $r: \mathcal{S} \times \mathcal{A} \mapsto \mathbb{R}$, an initial state distribution $p({s}_{t=0})$, a transition probability distribution $p\left({s}_{t+1} \mid {s}_{t},{a}_{t}\right)$, and a discrete-time finite or continuous-time infinite horizon $H$. MBRL methods sample ground truth data $\mathcal{D}_{i}=
\left\{\left({s}_{0}, {a}_{0}, {s}_{1}\right),\left({s}_{1}, {a}_{1}, {s}_{2}\right), \ldots\right\}$ from a specific task $\mathcal{M}_{i}$ and use this data to train a dynamics model $
p_{\theta}\left({s}_{t+1} \mid {s}_{t}, {a}_{t}\right)
$ that estimates the underlying dynamics of the task to approximate which state follows which action. This is done by optimizing the weights $\theta$ to maximize the log-likelihood of the observed data: 
\begin{equation}
\begin{aligned}
\theta^{*} &=\underset{\theta}{\operatorname{argmax}}\:{p}(\mathcal{D}_{i}^{\text {}}\mid \theta) \\
&=\underset{\theta}{\operatorname{argmax}} \sum_{\left({s}_{t}, {a}_{t}, {s}_{t+1}\right) \in {D}_{i}^{\text {}}} \log p_{\theta}\left({s}_{t+1} \mid {s}_{t}, {a}_{t}\right)
\end{aligned}
\end{equation}
The learned dynamics model is then utilized to optimize a sequence of actions (e.g., with model predictive control) or to optimize a policy \cite{Deisenroth2011}, \cite{Nagabandi2017}.
\subsection{Gradient-based Reinforcement Learning with REPTILE}
\label{GBML}
REPTILE from Nichol et al. \cite{Nichol} is a first-order implementation of MAML. During meta-training, task data in the form of $\mathcal{K}$ trajectories $\mathcal{D}_{i}^{\mathcal{}}=\left\{\left({s}_{1}, {a}_{1},{r}_{1} \ldots, {s}_{H}\right),\ldots, \mathcal{K} \right\}$ is sampled with roll-outs from $f_{\theta}$ or by taking random actions. A single task $\mathcal{M}_{i}$ is sampled from $p(M)$ without replacement for each training-iteration that passes the inner-level and outer-level once.  In the inner-level, task-specific parameters $\phi_i$ are generated by adapting $f_{\theta}$ with $g>1$ steps of stochastic gradient descent as $Alg$ and its learning rate $\psi = \beta$:
\begin{equation}
\phi_{i_{\text{REPTILE}}}= \theta-\nabla_{\theta}\mathcal{L}_{\mathcal{D}_{i}} (\theta_{})
\end{equation} 
In the outer-level the parameters $\theta$ are then being adjusted to minimize the euclidean distance between $\theta$ and $\phi_{i}$ with learning rate $\alpha$:
\begin{equation}
\theta \leftarrow \theta+\alpha \left(\phi_{i_{\text{REPTILE}}}-\theta\right)
\end{equation}
\begin{figure*}
  \includegraphics[width=\textwidth,height=5.2cm]{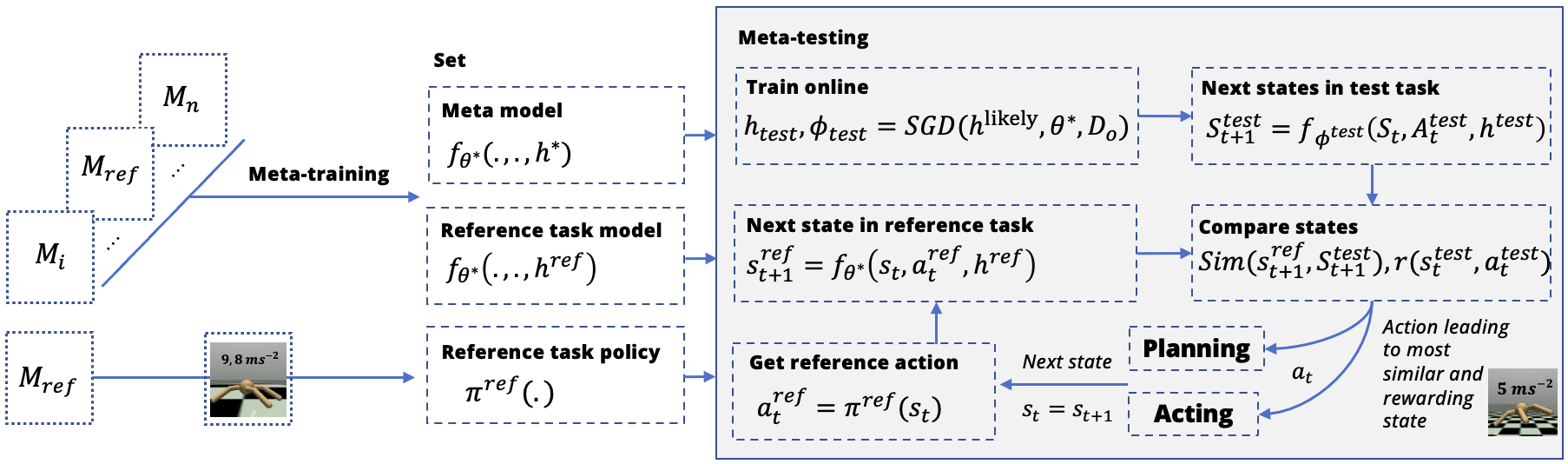}
  \caption{A high-level overview of the meta adaption controller. While being initialized with the most likely embedding, a trained meta-model is adapted to approximate future environment states and compare them to states of the reference task. Actions leading to states in the unseen task that are very similar to those reached by the RL policy in the reference task are then chosen to be executed in the environment.}
 \label{fig:mac}
\end{figure*}
\subsection{Model-based Meta-Reinforcement Learning using task embeddings}
Each dynamic encountered by an agent can be represented by an MDP and therefore interpreted as an RL task $\mathcal{M}_{i}$. Since, in real-world applications, new dynamics can appear at any time (e.g., a malfunctioning robot leg), a new task could appear at any time. Hence, a task can be understood as an arbitrary trajectory segment of $k$ timesteps under a specific dynamic. A meta-learner a meta-learner is trained to adapt to the distribution of these temporal fragments based on $o$ recent observations \cite{Nagabandi2018}, \cite{Kaushik2020}. FAMLE by Kaushik et al. \cite{Kaushik2020} extends a dynamics model input with an additional input $h$, which is a d-dimensional vector describing a task $\mathcal{M_\text{i}}$. By meta-training model parameters $\theta$ and embeddings $h$ jointly, several initial sets of model parameters are found, each conditioned on a task represented by $h_i$ resulting in a task conditioned dynamics model $p_{\theta}\left(s_{t+1} \mid s_{t}, a_{t}, h\right)$. If an unseen task $M_j$ appears, its similarity to prior tasks is measured. The most-likely task embedding $h_\text{likely}$ is then used to condition the model parameters and enable faster adaption.\newline
The meta-training process is described in Algorithm \ref{alg:ENN_train}. Prior to meta-training, $n$ tasks are sampled inside a simulation from $p(\mathcal{M})$ resulting in a set of meta-training tasks $\mathcal{M}$. For each $\mathcal{M}_i$ training data $\mathcal{D}_{i}=\left\{\left(s_{t}, a_{t}, s_{t+1}\right) \mid t=1, \ldots, N\right\}$ is sampled by a simulated robot randomly taking $N$ actions. Then, to be learned task embeddings $\mathbb{H}=\left\{h_{{i}} \mid i= 1\dots n \}\right.$corresponding to each task are initialized. During meta-training, relating to the meta-learning goal defined in Equation \ref{eqn:meta-learning-objective}, initial model parameters $\theta^\star$ and $n$ task embeddings $\mathbb{H}^\star$ are found that minimize the loss for any task $M$ sampled from $p(M)$:
\begin{equation}
\begin{array}{c}
\theta^{\star}, \mathbb{H}^\star=\arg \min _{\theta, {h}_{i}} \sum_{i=1}^{n} \mathcal{L}_{\mathcal{D}_{i}}(\phi_{i}, {h}_{i})\\
\text { where } {{\phi_{i},h_i}}=Alg_{\mathcal{D}_{i}}^{\psi}\left(\theta, h\right)
\end{array}
\end{equation}The loss of a task-conditioned dynamical model for a specific task $ \mathcal{D}_i \sim \mathcal{M}_i  $  be as follows: \begin{equation}\mathcal{L}_{\mathcal{D}_{i}}(\phi_{i}, h_{i})=-\frac{1}{K} \sum_{t=1}^{t=K} \log p_{\phi_{i}}({s}_{t+1} \mid {s}_{t},{a}_{t}, h_i) \end{equation}
Following the REPTILE algorithm, bi-level optimization is achieved by making a gradient-based, task-specific update of $\theta$ and $h$ in the inner level with a fixed $\psi$: 
\begin{equation}
\begin{split}
\phi_{i} = \theta-\nabla_{\theta}\mathcal{L}_{\mathcal{D}_{i}} (\theta, h)\\\
h'_i = h-\nabla_{h}\mathcal{L}_{\mathcal{D}_{i}}(\theta, h)
\end{split}
\end{equation}
and simultaneously updating $\theta$ and $h$ towards their task-specific counterparts in the outer level:
\begin{equation}
\begin{split}
\theta \leftarrow \theta+\alpha (\phi_{i_{\text{}}}-\theta)\\
h_i \leftarrow h_i+\alpha (h'_{i_{\text{}}}-h_i)
\end{split}
\end{equation}

\begin{algorithm}[H]
	\begin{algorithmic}[1]
		\REQUIRE Distribution $p(\mathcal{M})$ over tasks
		\REQUIRE Learning rate outer-level $\alpha \in \mathbb{R}^+$
		\REQUIRE Learning rate inner-level $\beta \in \mathbb{R}^+$
		\REQUIRE Number of sampled tasks $n$
		\REQUIRE Empty Dataset $\mathcal{D}_{}^{\text{meta-train}}=\{\}$
		\REQUIRE $Alg^{\psi}_{D_i}()$ as $k$ steps of stochastic gradient descent
		\FOR{$i=1 \dots n$}
			\STATE Sample a training task $\mathcal{M}_{i}$ from $p(\mathcal{M})$
			\STATE Save the task: $\mathbb{M} \leftarrow \mathcal{M}_{i}$
			\STATE Collect task data: $\mathcal{D}_{i} = \{(s_t, a_t, s_{t+1}) | t=1,\ldots,N\}$
			\STATE Save task data: $\mathcal{D}_{}^{\text{meta-train}} \leftarrow \mathcal{D}^{\text{meta-train}} \cup \{\mathcal{D_{\text{i}}}\}$
		\ENDFOR
		\FOR {$x = 0,1,...$}
			\STATE Sample task data: $\mathcal{D}_{i} \sim \mathcal{D}_{}^{\text{meta-train}}$
			\STATE Perform $k$ steps of SGD with: $\phi_i, h'_{_i} = Alg_{i}^{\psi=\beta}(\theta, h_{i})$
			\STATE Perform update: $\theta \leftarrow \theta + \alpha (\phi_i- \theta)$
			\STATE Perform update: $h_{i} \leftarrow h_{i} + \alpha(h'_{i} - h_{i})$
		\ENDFOR
		\RETURN{(${{\theta}}$, ${h}$) as (${{\theta}^{\star}}$, ${h}^{\star}$)}
	\end{algorithmic}
\caption{Meta-training process using REPTILE and an Embedding Neural Network}
\label{alg:ENN_train}
\end{algorithm}

During online adaptation (i.e. meta-testing) the dynamics model is adapted based on $o$ recent observations while making the assumption that a new task is taking place after every $k$ control steps. First, based on $o$ recent observations $\mathcal{D}_{j}=\left\{\left(s_{t}, a_{t}, s_{t+1}\right) \mid t=1, \ldots, o\right\}$ the most likely situational embedding $h_{Likely}$ is defined:
\begin{equation}
\label{eqn:most_likely_emb}
h_{\text {Likely }}=\arg \max _{h \in \mathbb{H}^{\star}} \mathbb{E}_{\mathcal{D}_{j}}\left[\log p_{\theta^{\star}}\left(s_{t+1} \mid s_{t}, a_{t}, h\right)\right]
\end{equation}Next, the dynamical model is updated online by simultaneously updating $h_{Likely}$ and $\theta^{\star}$ taking $g$ gradient steps:
\begin{equation}
\label{eqn:adapted_enn}
\begin{split}
\theta \leftarrow \theta-\beta \nabla_{\theta}\mathcal{L}_{D_j}(\theta, h_{\text{likely}})\\
h_j \leftarrow h_{likely}-\beta \nabla_{h}\mathcal{L}_{D_j}(\theta, h_{\text{likely}})
\end{split}
\end{equation}
\section{Agents forming adapted behavior through meta-learning}
Figure \ref{ant_anchor} displays a robot incentivized to walk in one specific direction as fast as possible. The meta-learning objective is to walk successfully in different gravitational settings. First, as in algorithm  \ref{alg:ENN_train}, data is collected by randomly taking actions in different gravitational settings. Next, a meta-learning method (e.g., REPTILE) is used to train a meta-model. While the robot achieves good performance during meta-testing with MPC, its adapted behavior in low gravitational settings is to jump and roll since this results in the highest reward (Fig. \ref{ant_roll}). In a real-world setting, similar adverse behaviors could have unknown consequences like damage to the robot or its environment. Designing a reward function that enables intended adaption is not a promising approach. First, developing the proper function for one specific task takes many trials, which is time-consuming.
Moreover, finding a reward function that works across various tasks is difficult. For example, settings with low gravity require constraining the motion of the robot not to jump or roll, whereas high gravity settings demand rotation flexibility. More complex meta-learning tasks are even more challenging.

\section{Apply preferred behavior to similar tasks}

Instead of designing a reward function that provides the right incentives for different tasks, a technique is needed that guarantees correct motion with minimal supervision. One possible solution is to add constraints to the MPC optimization problem. These constraints force the states predicted by the model to be similar to the predefined states that we call task-anchors $s^{anch} $. An example is shown in Fig. \ref{ant_anchor} (red arrows), where the task-anchor accounts for the robot's rotational motion so that the robot adapts to low gravity conditions without jumping or rolling.
\begin{alignat}{3}
\max_{a_{[\cdot]}} \quad& \sum_{t=0}^{H-1} r(s_{t}, a_{t}) &&          & \\
\text{s.t.: } \quad& {s}_{t+1} = f_{\phi^{test}}(s_{t}, a_{t}, h^{test}) &\quad \forall t=[1,H-1]\\ 
                    \quad& \lvert {s}_{t+1}  - {s}^{anch}_{t+1} \rvert \leq \delta &\quad \forall t=[1,H-1] 
\end{alignat}
Here $\delta$ is the similarity threshold.

Instead of finding the proper movement across all tasks, we only choose a movement of one specific task that may work well in similar tasks. This movement can be extracted from learned RL policies or classical feedback controllers. We call this task a reference task and the used policy (controller) a reference policy.
Our algorithm aims to find actions the robot has to take in a new test task to reach a similar outcome as the desired outcome in the reference task using the reference policy. 
To achieve that, our algorithm utilizes two variations of a meta-trained embedding neural network (ENN). The first variation entails the meta-trained network with meta parameters $\theta^{*}$ and learned embeddings $h^{*}$. The second variation entails the meta-trained network with meta parameters $\theta^{*}$ conditioned on the embedding of a reference task $h^{ref}$.\newline 
Putting these pieces together, MAC (Fig. \ref{fig:mac}) optimizes a sequence of states $s_{[\cdot]}$ and actions $a_{[\cdot]}$ to maximize the predicted reward in the test-task while also eventually ensuring dynamics feasibility:
\begin{alignat}{3}
\max_{ s_{[\cdot]}, a_{[\cdot]}} \quad& \sum_{t=0}^{H-1} r(s_{t}, a_{t})\label{eqn:mac}\\
\text{s.t.: } \quad& {s}_{t+1} = f_{\phi^{test}}(s_{t}, a_{t}, h^{test}) &\quad \forall t=[1,H-1]\\ 
                    \quad& \lvert {s}_{t+1}  - {s}^{ref}_{t+1} \rvert \leq \delta &\quad \forall t=[1,H-1]\label{eqn:constraint}
\end{alignat}
Where ${s}^{ref}_{t+1}$ is the desired outcome of the reference task given the current state ${s}_{t}$ and using the reference policy:
\begin{equation}
{s}^{ref}_{t+1}  = f_{\theta^{*}}(s_{t}, \pi^{ref}(s_{t}), h^{ref}) 
\end{equation}
The constraint \ref{eqn:constraint} is approximated using a similarity measurement between the predicted states and the reference state $Sim(s_{t+1}, s_{t+1}^{ref})$. As the similarity measure, we use the cosine similarity of the state vectors.\newline
\begin{figure}
\centerline{\includegraphics[scale=0.12]{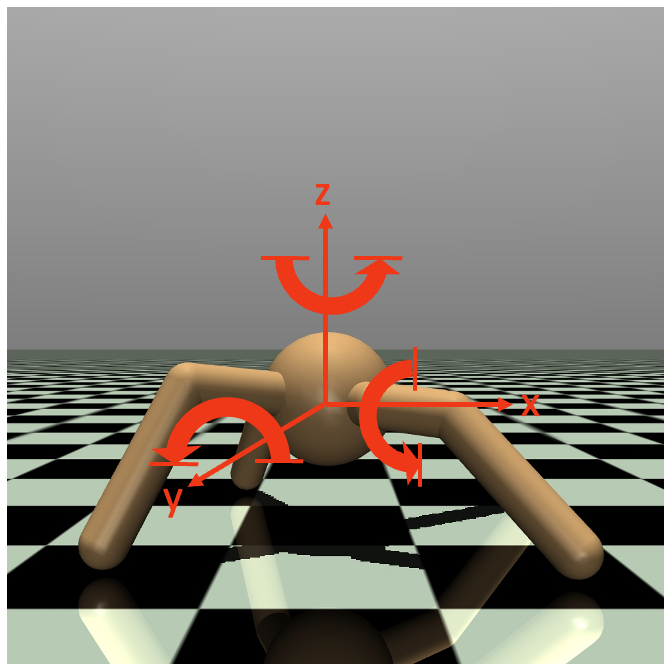}}
\caption{An ant robot incentivized to walk as fast as possible. The red arrows depict an anchor that restricts the robot in its rotation}
\label{ant_anchor}
\end{figure}
\begin{algorithm}[H]
\begin{algorithmic}[1]
 \REQUIRE Meta-learned parameters $\theta^{\star}$ and embeddings $\mathbb{H}^{\star}$
 \REQUIRE Meta Adaption Controller \textbf{MAC}$()$
 \REQUIRE Empty set of $o$ recent observations $\mathcal{D}_{o}=\{\}$
\WHILE {task not solved}
\STATE Determine most likely embedding $h_{Likely} \in \mathbb{H}^{\star}$ given $\mathcal{D}_{o}$ and $\theta^{\star}$ \hfill\COMMENT{see Eq. \ref{eqn:most_likely_emb}}
\STATE Execute $g$ steps of SGD using $\mathcal{D}_{o},\theta^{\star}, h_{Likely}$ and receive $\phi^{test}, h^{test}$ \hfill\COMMENT{see Eq. \ref{eqn:adapted_enn}}
\STATE Execute action $a_t = \textbf{MAC}(\phi^{test}, h^{test},s_t)$ and receive state $s_{t+1}$ \hfill\COMMENT{see Alg. \ref{alg:MAC_test}}
\STATE Save observation $\mathcal{D}_{o} \leftarrow \mathcal{D}_{o} \cup \{(s_t, a_t, s_{t+1})\}$
\STATE \textbf{if } $size(\mathcal{D}_{o}) > o$ \textbf{then} remove oldest observation from $\mathcal{D}_{o}$
\ENDWHILE
\end{algorithmic}
\caption{Meta-testing an Embedding Neural Network to adapt online using our control algorithm}
\label{alg:general_test}
\end{algorithm}
After meta-training an ENN with algorithm \ref{alg:ENN_train}, meta-testing (i.e., online adaption) is executed with algorithm \ref{alg:general_test}. Here the most likely embedding is determined, and the ENN is adapted to approximate future environment states. Subsequently, MAC is used to find the action to take in a new task that reaches a similar outcome as in the reference task.\newline
The MAC procedure is shown in Algorithm \ref{alg:MAC_test}, and can be summarized with the following steps:
\begin{enumerate}
    \item Given the current state $s_t$, sample a set of reference actions $A^{ref}_{t}$   using the reference policy $\pi^{ref}(s_t)$.
    \item Using the embedding of the reference task $h^{ref}$, estimate a set of reference states $S^{ref}_{t+1}$ that contains information about what states would follow if the reference actions in the reference task were executed.
    \item Sample a set of actions $A_t$ from  an unconditional Gaussian distribution $\mathcal{N}\left(\mu,\Sigma\right)$. These actions will be used together with the test task embedding $h^{test}$ to predict the following states $S_{t+1}$.
    \item Use the reward function $r(S_{t},A_{t})$ to estimate a set of rewards $R_t$ and measure the state similarity of the predicted states to the reference state ${Sim}(S_{t+1},S^{ref}_{t+1})$.
    \item From the sets $S_{t+1}$ and $A_{t+1}$ store only the most similar states and the actions used to reach these states.
    \item Repeat the procedure according to the planning horizon. 
    \item Choose the first action $a_t$ of the most similar action sequence with the highest reward.
    \item Update the distribution $\mathcal{N}\left(\mu,\Sigma\right)$  towards action sequences with higher similarity and reward.
\end{enumerate}

\begin{algorithm}[H]
\begin{algorithmic}[1]
\REQUIRE Reward function $r()$ ; similarity measure $Sim()$
\REQUIRE Action elites $\epsilon$ ; planning horizon $\eta$
\REQUIRE Reference policy $\pi^{ref}()$ 
\REQUIRE Task-specific configuration $\phi^{test}, h^{test}$
\REQUIRE Reference task configuration $\theta^{ref}, h^{ref}$
\REQUIRE Set of initial states $S_{t}$
\REQUIRE Empty set of next states $\mathcal{S}=\{\}$
\REQUIRE Empty set of actions to next states $\mathcal{A}=\{\}$
\REQUIRE Empty set of calculated rewards $\mathcal{R}=\{\}$
    \STATE Use random distribution $\mathcal{N}\left(\mu,\Sigma\right)$ to sample actions
    \FOR{$t <= \eta$}
        \STATE Sample set of ref. actions $A^{ref}_{t}$ with $\pi^{ref}(S_t)$
        \STATE Calculate ref. states $S^{ref}_{t+1}$ with $f_{\theta^{*}}(S_{t}, A^{ref}_{t}, h^{ref})$ 
        \STATE Sample set of actions $\mathcal{A}_{t}$ from $\mathcal{N}\left(\mu, \Sigma\right)$
        \STATE Get set of predictions of next states $S_{t+1}$ with $f_{\theta^{test}}(S_{t}, A_{t}, h^{test})$
        \STATE Calculate rewards $R_t$ with $r(S_{t},A_{t})$
        \STATE Calculate state similarity $S_{sim}$ with ${Sim}(S_{t+1},S^{ref}_{t+1})$
        \STATE Update $A_{t}$ based on $S_{sim}$ and $R$ to consists of the $\epsilon$ most similar states with the highest rewards
        \STATE $\mathcal{S}_{} \leftarrow \mathcal{S}_{} \cup \{S_{t+1}\}$
        \STATE $\mathcal{R}_{} \leftarrow \mathcal{S}_{} \cup \{R_{t}\}$
        \STATE $\mathcal{A}_{} \leftarrow \mathcal{S}_{} \cup \{A_{t}\}$
    \ENDFOR
\STATE Extract the first action $a_{t}$ of the most similar action sequence with the highest reward.
\STATE Update $\mathcal{N}\left(\mu, \Sigma\right)$ based on $\mathcal{A}$
\RETURN{$a_{t}$}
\end{algorithmic}
\caption{Meta Adaption Controller (MAC)}
\label{alg:MAC_test}
\end{algorithm}
\section{Experiments}
We compare our meta adaption controller (MAC) algorithm with two baseline algorithms. The algorithms are compared during meta-testing within four different environments where the agent must quickly adapt to new tasks and collect as much reward as possible (Figure \ref{figenvs}). The environments are based on the MuJoCo physics engine developed by Todorov et al. \cite{todorov2012mujoco} and among them previous meta-RL literature \cite{Nagabandi2018}: Halfcheetah-disabled, Halfcheetah-pier, Ant-disabled, Ant-gravity.\newline
Each baseline uses a multilayer perceptron (MLP) with three hidden layers, each consisting of 512 neurons. The first baseline (RMPC) trains the MLP on the MAML first-order implementation REPTILE by Nichol et al. \cite{Nichol} and uses MPC for meta-testing. By following the approach of FAMLE by Kaushik et al. \cite{Kaushik2020}, the second baseline (FMPC) extends the MLP with an additional embedding input, meta-trains the resulting embedding neural network (ENN) using REPTILE, and uses MPC for meta-testing. To extract reference actions, MAC uses a policy from an actor-critic module trained on one reference task per environment with the soft actor-critic algorithm by Haarnoja et al. \cite{Haarnoja2018}. The reference task in each environment corresponds to the goal of its base environment not using any meta-task (i.e., Ant without disability and in standard gravitational setting; HalfCheetah without disability). Further, MAC reuses the trained ENN as explained in algorithm \ref{alg:general_test}.\newline
A hyperparameter search regarding meta-training and meta-testing was carried out for each environment algorithm combination. To compare the individual performances in each environment, we sampled a new task after 500 steps. During testing, every task is sampled five times with different environment seeds. With each environment tested on five different seeds per task, our experiments show that the MAC algorithm outperforms both baselines with the exception when the robot jumps and rolls in the ant gravity environment as depicted in Figure \ref{ant_roll}. The experiment results are displayed in Table \ref{table}.
\begin{figure}
\centerline{\includegraphics[scale=0.15]{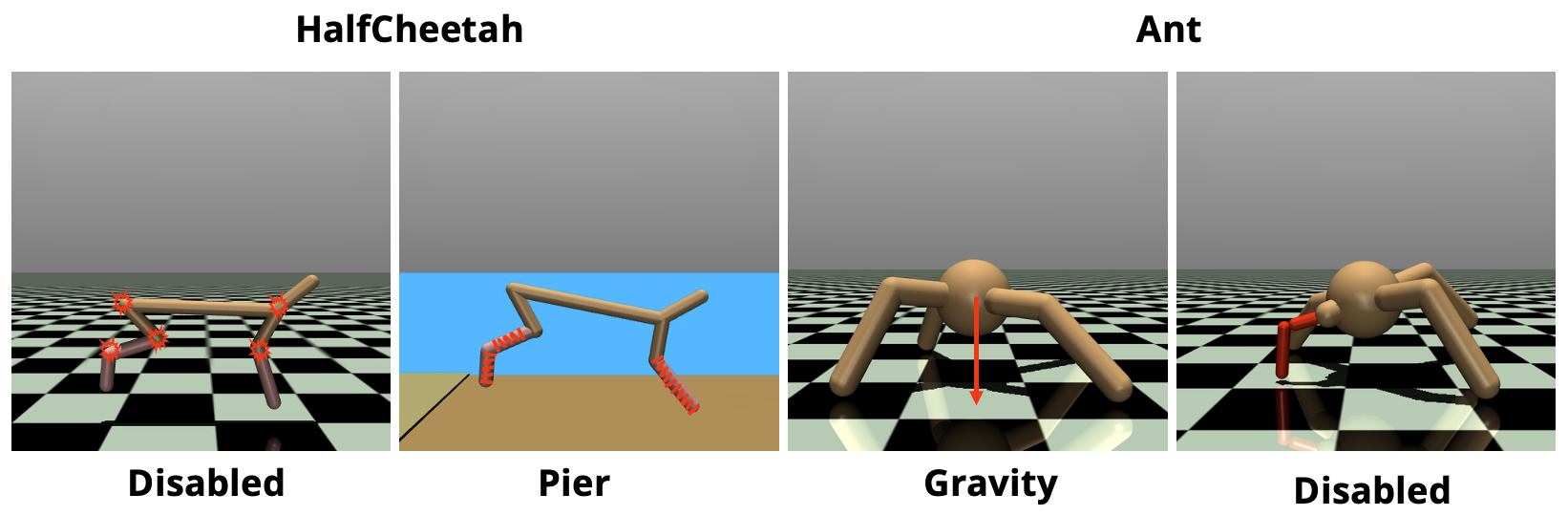}}
\caption{Environments used to test our algorithm. In each environment, the corresponding robot needs to run as fast as possible in one direction. Halfcheetah-disabled blocks different joints of a cheetah robot. Halfcheetah-pier changes the cheetah's limb flexibility while running on moving ground. Ant-disabled resizes different legs of an ant robot. In Ant-Gravity, the environment's gravitational setting is adjusted.}
\label{figenvs}
\end{figure}
\begin{table}
        \caption{Meta-testing in different environments}
        \begin{tabular}{ |p{1.9cm}||p{0.6cm}|p{0.7cm}|p{0.9cm}|p{1cm}|p{1cm}| }
            \hline
            Environment & $j$ tasks & env steps & RMPC & FMPC & MAC\\
            \hline
            Ant-gravity&8&20000&18840*&5981&10838\\
            Ant-disabled&1&2500&2001&1856&2397\\
            Hc-disabled&2&5000&2109&9762&10513\\
            Hc-pier&2&5000&2570&4342&4393\\
            \hline
            \multicolumn{6}{l}{*: Robot jumps and rolls in environment leading to high rewards}\\
        \end{tabular}
\label{table}
\end{table}
\section{Conclusion}
In this paper, we presented MAC, a robot control algorithm that employs meta-reinforcement learning to apply a preferred robot behavior from one task to many similar tasks. At its core is the combination of a meta-trained embedding neural network and a RL policy of a reference task. While adapting the neural network online, it predicts which actions need to be taken in unseen tasks to mimic the behavior of the reference task obtained by the policy. Our experiments demonstrated that this mechanism works across various tasks in different environments and outperforms meta-testing with model predictive control in different model-based meta-reinforcement learning setups. 
\bibliographystyle{IEEEtran}
\bibliography{bibliography.bib}
\end{document}